\documentclass[10pt,twocolumn,letterpaper]{article}

\usepackage[pagenumbers]{cvpr} % To force page numbers, e.g. for an arXiv version

\usepackage{colortbl}
\usepackage{multirow}

\usepackage[
  pagebackref,
  breaklinks,
  colorlinks,
  linkcolor=red,      % \ref（章节、图表等引用）为红色
  citecolor=blue,     % \cite（文献引用）为蓝色
  urlcolor=magenta    % 可选：超链接（网址）为洋红色
  ]{hyperref}

\definecolor{cvprblue}{rgb}{0.21,0.49,0.74}

\def\paperID{11463} % *** Enter the Paper ID here
\def\confName{CVPR}
\def\confYear{2026}

\newcommand{\vsection}[1]{%
\vspace{0pt}  % 在\section之前添加垂直空间
\section{#1}  % 原始的\section命令
\vspace{0pt}  % 在\section之后添加垂直空间
}

\newcommand{\vsubsection}[1]{%
\vspace{0pt}  % 在\section之前添加垂直空间
\subsection{#1}  % 原始的\section命令
\vspace{0pt}  % 在\section之后添加垂直空间
}

\title{From Events to Clarity: The Event-Guided Diffusion Framework for Dehazing}

\author{
Ling Wang\textsuperscript{*}, Yunfan Lu\textsuperscript{*},
Wenzong Ma, Huizai Yao, Pengteng Li, Hui Xiong\textsuperscript{\dag} \\
The Hong Kong University of Science and Technology (GuangZhou)\\
{\tt\small \{lwang851, ylu066, wma423, hyao032, pli807\}@connect.hkust-gz.edu.cn, xionghui@ust.hk} \\
[2pt]
{\small
\textsuperscript{*}Equal contribution.\quad
\textsuperscript{\dag}Corresponding author.
}
}

\begin{document}
\maketitle

\begin{abstract}
Clear imaging under hazy conditions is a critical task.
Prior-based and neural methods have improved results.
However, they operate on RGB frames, which suffer from limited dynamic range.
Therefore, dehazing remains ill-posed and can erase structure and illumination details.
To address this, we use event cameras for dehazing for the \textbf{first time}.
Event cameras offer much higher HDR ($120 dB~vs.~60 dB$) and microsecond latency, therefore they suit hazy scenes.
In practice, transferring HDR cues from events to frames is hard because real paired data are scarce.
To tackle this, we propose an event-guided diffusion model that utilizes the strong generative priors of diffusion models to reconstruct clear images from hazy inputs by effectively transferring HDR information from events.
Specifically, we design an event-guided module that maps sparse HDR event features, \textit{e.g.,} edges, corners, into the diffusion latent space.
This clear conditioning provides precise structural guidance during generation, improves visual realism, and reduces semantic drift.
For real-world evaluation, we collect a drone dataset in heavy haze (AQI = 341) with synchronized RGB and event sensors. Experiments on two benchmarks and our dataset achieve state-of-the-art results.
\url{https://evdehaze.github.io/}
\end{abstract}

\vsection{Introduction}
Haze degrades image quality by scattering and absorbing light, which leads to a significant reduction in the dynamic range of captured images.
This effect is especially pronounced for active pixel sensors ~\cite{zhou2023computational}, which already operate under limited dynamic range, resulting in severe loss of visual detail in hazy conditions.
As a result, the visibility of key textures and edges is greatly diminished, which in turn affects both the visual quality and the effectiveness of downstream computer vision tasks.
Therefore, recovering clear visual content under haze becomes particularly important for safety-critical applications such as photography, drone-based monitoring, and autonomous driving.

Earlier dehazing methods focus on atmospheric scattering models~\cite{mccartney1976optics,narasimhan2003contrast} and statistical priors~\cite{lu2022priors,he2010single}.
For instance, \cite{mccartney1976optics} proposes the atmospheric scattering model, while \cite{narasimhan2003contrast} analyzes contrast degradation.
Although these methods provide physical interpretability, they often fail to generalize well in complex real-world scenes due to their strong assumptions and handcrafted rules.
With the advent of deep learning methods \cite{cai2016dehazenet,zhang2018densely,li2017aod,qin2020ffa,an2022semi,liu2020end} introduce various forms of supervision and advanced network designs, achieving consistent improvements across synthetic datasets.
Recently, diffusion-based generative models \cite{ho2020denoising,lin2024diffbir,yu2023high} have shown strong potential.
However, these models still rely entirely on the RGB images captured by APS sensors, which inherently suffer from limited dynamic range.
As a result, existing methods remain limited under dense haze conditions due to irreversible information loss, often producing visible artifacts and unrealistic textures.

\begin{figure}
\centering
\includegraphics[width=0.99\linewidth]{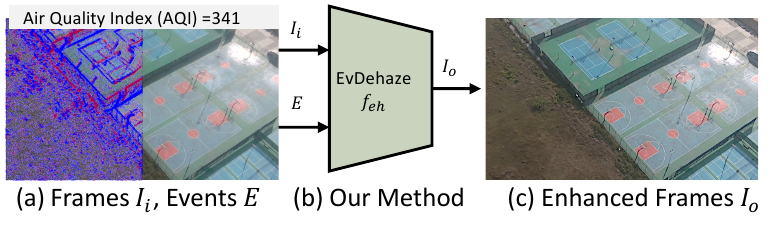}
\vspace{-7pt}
\caption{\small
Overview of the motivation and the proposed \textit{EvDehaze} method. Haze compresses the dynamic range of conventional RGB frames $I_i$ (see Sec.~\ref{sec:preliminary}) captured in heavy pollution (AQI = 341), as illustrated in \textbf{(a)}-right. In contrast, events $E$ captured by event cameras offer a much higher dynamic range (up to $120 dB$), as illustrated in \textbf{(a)}-left, and thus provide critical cues under haze. Our method, \textit{EvDehaze}, uses this events to guide dehazing (Sec.~\ref{sec:method}), producing clearer outputs $I_o$ as shown in \textbf{(c)}.
}
\vspace{-13pt}
\label{fig:0-CoverFigure}
\end{figure}

Event cameras, thanks to their asynchronous triggering mechanism, capture brightness changes at the microsecond scale with an exceptionally high dynamic range \cite{mueggler2017event}.
This makes them naturally capable of detecting fine lighting variations in scenes, providing critical advantages over conventional cameras in many imaging tasks.
Event sensors have been successfully applied in tasks such as low-light enhancement \cite{liang2024towards}, HDR imaging \cite{han2023hybrid}, video super-resolution~\cite{lu2023learning}, deblurring \cite{pan2019bringing}, and deraining~\cite{zhang2025egvd}.
These successes motivate us to explore whether the high dynamic range of events can be transferred to restore the visibility of images degraded by haze, as shown in Fig.~\ref{fig:0-CoverFigure}.
However, this goal remains challenging — especially for dehazing, where real-world paired supervision is unavailable.
Most prior approaches \cite{liang2023coherent,liang2024towards} with events-guidance rely on large-scale paired datasets to learn accurate mappings.
In hazy environments, however, collecting aligned hazy and clean image pairs is nearly impossible due to uncontrollable factors such as fog thickness and lighting changes.
This leads to a core research question: \textbf{How can we effectively inject the high dynamic range of events into a dehazing model without paired real-world dataset for supervision?}

To address this question, we reformulate event-based dehazing as a conditional image generation task, guided by sparse event data within a diffusion framework.
Diffusion models, known for their strong generative priors and robust generalization to real-world data, have achieved impressive performance in various image synthesis tasks~\cite{zhao2023uni,blattmann2023stable,liu2024sora,sepehri2024mediconfusion}.
This makes them particularly suitable when the supervision is unavailable or impractical.
Crucially, the high dynamic range and temporal precision of events offer strong structural cues that can resolve this ambiguity.
To exploit this, we propose an event-guided diffusion model that incorporates both structural and illumination information from sparse events directly into the generative process.
Rather than explicitly translating events into RGB images, we inject event features into the latent denoising steps, allowing the model to leverage event-based guidance implicitly.
This approach enables the generation of realistic dehazed images by aligning the sampling trajectory with physically consistent edge and contrast priors from events.

To validate the effectiveness of our framework, we conduct experiments on two public synthetic datasets and a newly collected real-world dataset.
This real-world dataset is collected via drone under real weather conditions (AQI = 341) for qualitative evaluation.
\textit{To the best of our knowledge, this is the \textbf{first} real-world dehazing dataset that provides paired RGB and events, establishing a valuable foundation for future research on event-based dehazing.}
Our method achieves state-of-the-art performance across all diffusion-based methods, confirming both its theoretical soundness and practical applicability.
In summary, we make three key contributions in this work:
\textbf{(1)} We first introduce and formulate \textit{event-guided image dehazing} as a conditional generation problem under a diffusion framework.
\textbf{(2)} We build the first real-world RGB-event dehazing dataset captured under heavy haze conditions for practical evaluation.
\textbf{(3)} We propose \textit{EvDehaze}, a diffusion-based model that leverages temporally precise and HDR event features for structural guidance, achieving state-of-the-art results among diffusion-based approaches.

\noindent\textbf{Clarification:} Our goal is not to surpass RGB-only restorers on PSNR or SSIM. \textit{EvDehaze} is a generative, event-guided diffusion method that targets {perceptual realism}: clearer edges, natural contrast, and fewer artifacts in real haze. These objectives differ from pixel-wise fidelity and are not directly comparable to supervised restoration.

\vsection{Related Works}

\textbf{Frame-based Dehazing:} Traditional image-based dehazing methods rely on the atmospheric scattering model (ASM) \cite{gui2023comprehensive}, and commonly use assumptions such as the dark channel prior \cite{he2010single} to recover clear images.
However, these methods have limitations in handling complex real-world scenes.
In recent years, deep learning has significantly advanced dehazing research, such as DehazeNet \cite{cai2016dehazenet}, PMHLD \cite{chen2020pmhld}, and the transmission propagation network \cite{liu2018learning} have achieved notable improvements.
Additionally, some approaches have completely abandoned the ASM framework and directly employed end-to-end supervised learning for dehazing \cite{sharma2020scale, zhang2021hierarchical}.
Nevertheless, these supervised methods heavily depend on large-scale paired synthetic datasets, which limits their effectiveness in real-world scenarios.
Therefore, recent research \cite{li2021you,mo2022dca} has moved toward unsupervised dehazing methods without the need for paired data.
\textit{Despite this progress, existing methods only rely on RGB frames, which have limited dynamic range and thus face significant challenges in dynamically scenes.}

\noindent\textbf{Event-based Image Enhancement:}
Event cameras have shown exciting potential in image reconstruction tasks.
Early works apply events across diverse imaging tasks, \textit{e.g.,} video frame interpolation~\cite{tulyakov2021time,ma2025timelens}, image deblurring~\cite{sun2022event,kim2024frequency,zhang2024neural,lu2024uniinr}, rolling shutter correction~\cite{zhou2022evunroll,erbach2023evshutter}, and HDR imaging \cite{messikommer2022multi}, overcoming the limitations of conventional cameras in challenging conditions~\cite{jiang2023event,yang2023learning}.
Beyond reconstruction, events help mitigate environmental degradations, \textit{e.g.,} removal of rain and snow~\cite{wang2023unsupervised,fu2024event,muglikar2025event}.
Zhang \textit{et al.}\cite{zhang2023egvd} propose a framework that leverages motion cues from events, while Fu \textit{et al.}\cite{fu2024event} introduce a heterogeneous network for deraining.
\textit{However, despite this progress, applying events to image dehazing remains an open question.}

\noindent\textbf{Diffusion Models:}
Diffusion models learn strong priors from large-scale data to push image quality beyond sensor limits.
Early methods denoise iteratively~\cite{ho2020denoising,song2020score}, while latent diffusion works in a compressed latent space to speed synthesis~\cite{rombach2022high}.
These models now support video synthesis~\cite{blattmann2023stable,blattmann2023align,liu2024sora} and personalized text-to-image~\cite{ruiz2023dreambooth,li2024cosmicman}.
Additionally, recent methods such as ControlNet~\cite{zhang2023adding}, MultiDiffusion~\cite{bar2023multidiffusion}, and MimicMotion~\cite{zhang2024mimicmotion} introduce fine-grained conditional control, enabling more flexible and guided generation.
Diffusion models have also been applied to real-world image restoration~\cite{yu2024scaling}, further demonstrating their practical effectiveness.
\textit{However, in ill-posed tasks like dehazing, they may still generate unrealistic textures or artifacts due to insufficient structural constraints.}

\vsection{Preliminaries\label{sec:preliminary}}

We first analyze how haze degrades image quality by compressing the dynamic range, and then explain the pivotal role of event guidance within the DDIM sampling process.

\noindent\textbf{HDR for Dehaze:~}
The dynamic range of an imaging system is the ratio between the maximum and minimum detectable luminance $\text{DR}_{\mathrm{true}} = J_{\max} / {J_{\min}}$, where $J_{\max}$ and $J_{\min}$ denote the highest and lowest scene radiance that can be recorded by the sensor.
In hazy environments, image formation obeys the atmospheric scattering model:
\begin{equation}\small~
I(x)=J(x)\,t(x)+A\bigl(1-t(x)\bigr),
\end{equation}
where $I(x)$ is observed intensity at pixel $x$, $J(x)$ is scene radiance at $x$, $t(x)\in[0,1]$ is transmission describing the portion of light that reaches the camera, and $A$ is global atmospheric light (air-light).
For analytical clarity, we assume a spatially constant transmission $t$.
Let
\begin{equation}\small~
a = J_{\min}\,t, \quad
b = A\,(1-t).
\end{equation}
The observed minimum and maximum intensities become
\begin{equation}\small~
I_{\min}=a+b,\qquad I_{\max}=k\,a+b,
\end{equation}
with $\text{DR}_{\mathrm{true}}=J_{\text{max}} / J_{\text{min}} = k >1$.
Hence the observed dynamic range is
\begin{equation}\small~
\text{DR}_{\mathrm{obs}}
   =\frac{I_{\max}}{I_{\min}}
   =\frac{k\,a+b}{a+b}
   < k=\text{DR}_{\mathrm{true}},
\end{equation}
which proves that haze \emph{compresses} the dynamic range. The closer $t$ is to zero (dense haze), the smaller $a$ becomes, and the more DR is lost.
Therefore, high-dynamic-range sensing is essential for retaining contrast and enabling effective dehazing under severe visibility degradation.

\noindent\textbf{DDPM and DDIM:~}
DDPM~\cite{ho2020denoising} samples \(x_{0} \sim p_{data }(x)\) from the data distribution. Gaussian noise is then injected to the sampled \(x_{0}\) up to \(T\) timesteps, finally get a standard Gaussian distribution \(x_{T} \sim \mathcal{N}(0, I)\). The simplified forward process follows:
\begin{equation}\small~
q\left(x_{t} | x_{0}\right)=\mathcal{N}\left(x_{t} ; \sqrt{\alpha_{t}} x_{0},\left(1-\alpha_{t}\right) I\right),
\end{equation}
where $\bar{\alpha}t = \prod{s=1}^t (1 - \beta_s)$ denotes the cumulative product of the variance schedule ${\beta_s}$.
To recover data from noise, a denoiser $\epsilon_\phi$ is trained to predict noise through reweighted Evidence Lower Bound (ELBO) minimizing optimization, which follows:
\begin{equation}\small~
\mathcal{L}_{simple }=\mathbb{E}_{x_{0}, t, \epsilon\sim\mathcal{N}(0, I)}\left[\left\| \epsilon - \epsilon_{\phi}\left(x_{t}, t\right)\right\| ^{2}\right].
\label{eq:ELBO}
\end{equation}
Here, $x_t$ is derived via $x_t = \sqrt{\alpha_t}x_0 + \sqrt{1-\alpha_t}\epsilon$, enabling direct noise prediction without explicit likelihood computation and \(\epsilon_{\phi}(x_{t}, t)\) is the noise predicted by the denoiser.
The reverse process reconstructs data through sampling as shown in Eq.~\ref{eq:q_e}, where $\mu_\theta$ and $\Sigma_\theta$ denote the mean and covariance learned by the model.
\begin{equation}\small~
\label{eq:q_e}
p_{\theta}\left(x_{t - 1} | x_{t}\right)=\mathcal{N}\left(x_{t - 1} ; \mu_{\theta}\left(x_{t}, t\right), \Sigma_{\theta}\left(x_{t}, t\right)\right)
\end{equation}
Despite its effectiveness, DDPM suffers from high sampling latency due to its sequential and Markovian nature. To address this, DDIM (Denoising Diffusion Implicit Models)\cite{song2020denoising} defines a non-Markovian deterministic sampling procedure that shares the same training objective as Eq.~\ref{eq:ELBO}, but accelerates inference. The DDIM reverse process is defined as Eq.~\ref{eq:Markovian-diffusion-process}, where $\sigma_t^2$ is a user-defined variance parameter that allows trade-off between stochasticity and determinism.
\begin{equation}\small~
\label{eq:Markovian-diffusion-process}
q_{\sigma}\left(x_{t-1} | x_{t}, x_{0}\right)=\mathcal{N}\left(x_{t-1} ; \tilde{\mu}_{t}\left(x_{t}, x_{0}\right), \sigma_{t}^{2} I\right)
\end{equation}
The actual DDIM sampling update can be written as Eq.~\ref{eq:latst-eq}, where $z \sim \mathcal{N}(0, I)$ and $\sigma_t$ is a hyperparameter controlling the diffusion noise scale.
\begin{equation}\small~
\label{eq:latst-eq}
\begin{aligned}
x_{t-1} = \sqrt { \alpha _{t-1}} \left ( {\left(x_t - \sqrt {1- \bar { \alpha }_t} \cdot \epsilon _ \phi (x_t, t)\right)}/{ \sqrt { \bar { \alpha }_t}} \right ) \\+\sqrt {1- \alpha _{t-1}- \sigma _t^2} \cdot \epsilon _ \phi (x_t, t) + \sigma _t z
\end{aligned}
\end{equation}

\vsection{Methods\label{sec:method}}

\begin{figure*}
\centering
\includegraphics[width=0.95\linewidth]{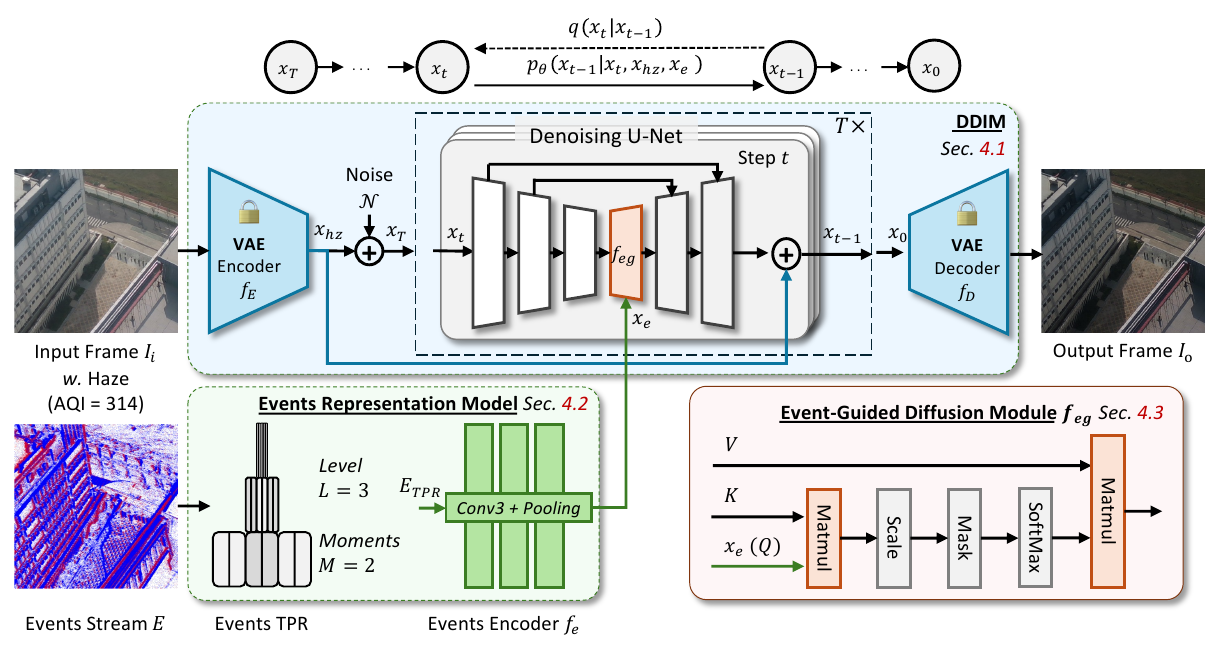}
\vspace{-10pt}
\caption{\small\small Overview of the proposed \textit{EvDehaze} framework. Given a hazy frame $I_i$ and its corresponding event stream $E$, our model generates a dehazed output $I_o$. The pipeline consists of three main components: (1) a frozen VQ-VAE~\cite{razavi2019generating} for encoding and decoding image latents; (2) an Event Encoder that extracts multi-scale representations from $E$ via TPR and Conv+Pooling; and (3) a denoising U-Net that performs iterative refinement from noise $x_T$ to clean latent $x_0$, guided by event features via cross-attention.\label{fig:3-Methods}}
\vspace{-10pt}
\end{figure*}

\begin{figure}
\centering
\includegraphics[width=0.99\linewidth]{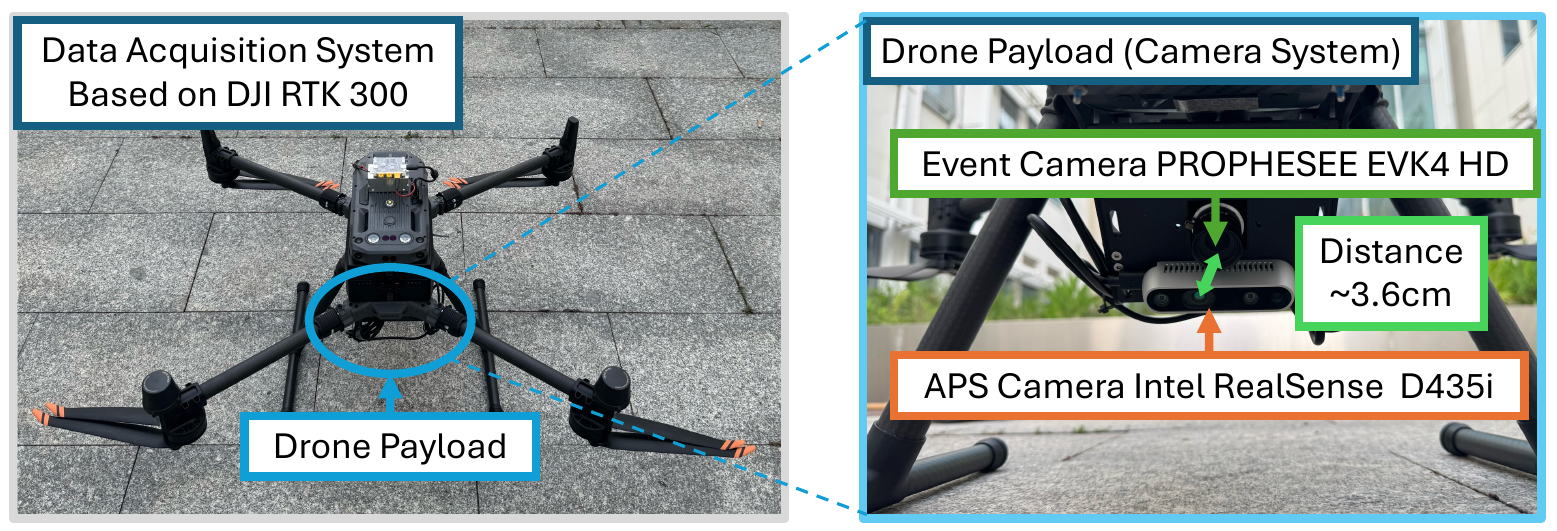}
\vspace{-4pt}
\caption{\small Our real-world data acquisition system mounted on the DJI Matrice 300 RTK platform. The payload includes a PROPHESSEE EVK4 HD event camera and an Intel RealSense D435i RGB camera with closely aligned viewpoints. The setup enables synchronized RGB-event recording in outdoor hazy conditions for long-range imaging analysis.}
\label{1-RealDataset}
\end{figure}

Our objective is to recover high-quality images from hazy inputs by leveraging the high dynamic range cues provided by event data.
To this end, we propose a novel event-guided diffusion framework, \textbf{\textit{EvDehaze}}, denoted as $f_{\text{edh}}$, as illustrated in Fig.~\ref{fig:3-Methods}.
Given a hazy RGB frame $I_i$ and the corresponding event stream $E$, our model predicts a dehazed output frame $I_o = f_{\text{edh}}(I_i, E)$.

The overall pipeline consists of three main components:
(1) \textbf{DDIM backbone with latent VQ-VAE:} A frozen VQ-VAE~\cite{razavi2019generating} first encodes the hazy frame into a latent representation $x_{hz} = f_E(I_i)$, where $f_E$ is the encoder. The latent is then injected with Gaussian noise to form $x_T$, and a DDIM-based denoising U-Net iteratively generates a clean latent $x_0$, which is finally decoded to the output image $I_o = f_D(x_0)$ by the frozen VQ-VAE decoder $f_D$.
(2) \textbf{Events Representation Model $f_e$:} The raw event stream $E$ is encoded into a multi-scale, high dynamic range feature $x_e = f_e(E)$.
This is achieved via the event temporal pyramid representation~\cite{lu2024hr} followed by a convolutional encoder with pooling layers.
(3) \textbf{Event-Guided Diffusion Module $f_{eg}$:} During each DDIM denoising step, we inject the event features $x_e$ into intermediate U-Net layers via cross-attention. This ensures that the generative trajectory is guided by physically grounded edge and contrast cues from events feature, enabling realistic, artifact-free dehazing.

\vsubsection{DDIM Backbone with Latent VQ-VAE}

Our framework builds upon a DDIM~\cite{song2020denoising} denoising process operating in the latent space of a frozen VQ-VAE~\cite{razavi2019generating}.
As shown in Fig.~\ref{fig:3-Methods}, the input hazy frame $I_i$ is first encoded by the VQ-VAE encoder $f_E$ into a compact latent representation $x_{hz} = f_E(I_i)$.
To simulate the forward diffusion process, we add Gaussian noise $\mathcal{N} \sim \mathcal{N}(0, I)$ to obtain the noisy latent:
\begin{equation}\small~
x_T = x_{hz} + \mathcal{N}.
\end{equation}

The reverse process then proceeds for $T$ steps, gradually refining $x_t$ into a clean latent $x_0$ using a DDIM sampler. At each step $t$, a denoising U-Net predicts $x_{t-1}$ by conditioning on the noisy input $x_t$, the original latent $x_{hz}$, and the external event features $x_e$:
\begin{equation}\small~
p_\theta(x_{t-1} \mid x_t, x_{hz}, x_e).
\end{equation}
For the mathematical formulation of DDPM and DDIM, please refer to Sec.~\ref{sec:preliminary}. This design allows the generation trajectory to remain anchored around the content of $I_i$, while being guided by the high dynamic range cues from the event feature $x_e$.
Once the clean latent $x_0$ is recovered, it is passed through the frozen VQ-VAE decoder $f_D$ to reconstruct the dehazed frame:
\begin{equation}\small~
I_o = f_D(x_0).
\end{equation}
By operating in a low-dimensional latent space, this architecture greatly improves inference efficiency and memory usage, while benefiting from the strong generative priors learned by the VQ-VAE and the structured noise modeling of the diffusion process.

\begin{table*}[t!]
\setlength{\tabcolsep}{13.6pt}
\renewcommand{\arraystretch}{1.0}
\centering
\small
\caption{\small \textbf{Quantitative Results.} Performance comparison of various state-of-the-art dehazing methods on both synthetic (SOTS) and real-world (NH-HAZE) datasets. The best overall results are shown in \textbf{\textit{italic}}, while the second-best are \underline{underlined}. \textcolor{red}{Red} indicates the highest performance among diffusion-based methods, and \textcolor{blue}{blue} highlights the second-best within the diffusion category.
}
\vspace{-6pt}
\label{tab:dehazing_comparison}
\resizebox{\linewidth}{!}{
\begin{tabular}{cl|ccc|ccc|r}
\toprule[0.5mm]
\multirow{2}{*}{\textbf{Diffusion}} & \multirow{2}{*}{\textbf{Method}} & \multicolumn{3}{c|}{\textbf{SOTS}} & \multicolumn{3}{c|}{\textbf{NH-HAZE}} & \multirow{2}{*}{\textbf{Params}} \\
 & & PSNR$\uparrow$ & SSIM$\uparrow$ & LPIPS$\downarrow$ & PSNR$\uparrow$ & SSIM$\uparrow$ & LPIPS$\downarrow$ & \\
\midrule[0.5mm]
$\times$          & DCP~\cite{he2010single}            & 15.09 & 0.765 & 0.069 & 10.57 & 0.520 & 0.399 & - \\
$\times$          & DehazeNet~\cite{cai2016dehazenet}  & 20.64 & 0.800 & 0.242 & 16.62 & 0.524 & 0.529 & \underline{0.01M} \\
$\times$          & AOD-Net~\cite{li2017aod}           & 19.82 & 0.818 & 0.099 & 15.40 & 0.569 & 0.495 & \textit{0.002M} \\
$\times$          & FFA-Net~\cite{qin2020ffa}          & 36.39 & \underline{0.989} & \textbf{\textit{0.005}} & \underline{19.87} & \textbf{\textit{0.692}} & 0.365 & 4.68M \\
$\times$          & MPRNet~\cite{mehri2021mprnet}      & 32.14 & 0.983 & 0.011 & 17.88 & 0.631 & 0.368 & 15.74M \\
\hline
$\times$  & SwinIR~\cite{liang2021swinir}      & 24.93 & 0.932 & 0.049 & 16.15 & 0.623 & 0.479 & 0.91M \\
$\times$  & Restormer~\cite{zamir2022restormer}& \underline{38.43} & \underline{0.989} & \underline{0.009} & 18.32 & 0.635 & 0.355 & 26.13M \\
$\times$    & Dehamer~\cite{guo2022image}        & 36.63 & 0.988 & \textbf{\textit{0.005}} & \textbf{\textit{20.66}} & \underline{0.684} & \textbf{\textit{0.230}} & 132.50M \\
$\times$    & \textbf{Restormer+EGDM}\textit{\small(ours)} & \textbf{\textit{39.12}}  & \textbf{\textit{0.990}} & \underline{0.009} & 19.23 & 0.676 & 0.331 & 32.86M \\
\hline
\rowcolor{gray!10}
\checkmark    & IR-SDE~\cite{luo2023image}      & \textcolor{blue}{33.82} & \textcolor{blue}{0.984} & \textcolor{blue}{0.014} & 12.59 & 0.520 & 0.361 & 537.21M \\
\rowcolor{gray!10}
\checkmark    & ResShift~\cite{yue2023resshift} & 29.06 & 0.950 & 0.017 & \textcolor{blue}{16.26} & \textcolor{blue}{0.625} & \textcolor{blue}{0.327} & \textcolor{red}{114.65M} \\
\specialrule{0.5mm}{0pt}{0pt} 
\rowcolor{gray!10}
\checkmark    & \textbf{EvDehaze}\textit{\small(ours)}           & \textcolor{red}{34.12} & \textcolor{red}{0.986} & \textcolor{red}{0.012} & \textcolor{red}{18.43} & \textcolor{red}{0.637} & \underline{\textcolor{red}{0.313}} & \textcolor{blue}{122.68M} \\
\specialrule{0.5mm}{0pt}{0pt} 
\end{tabular}
}
\vspace{-10pt}
\end{table*}

\vsubsection{Events Representation Model \label{sec:events-guidance}}

To effectively guide the denoising trajectory of the diffusion process, the Events Representation Model aims to extract rich structural cues from the high dynamic range and high temporal resolution of the input event stream $E$. Inspired by prior works on multi-scale temporal encoding, we adopt a Temporal Pyramid Representation (TPR)~\cite{lu2024hr} to construct a compact yet expressive event feature map $x_e = f_e(E)$, as shown in Fig.~\ref{fig:3-Methods}.

Unlike conventional images, events naturally respond to local brightness changes, providing precise edge and motion information that is resilient to haze. However, directly modeling this fine-grained temporal structure poses computational challenges. To balance resolution and efficiency, we build a TPR by dividing $E$ into $L=3$ hierarchical levels, each covering progressively smaller temporal windows. Within each level $l \in \{1,2,3\}$, we further partition the events into $M=2$ time bins, forming a voxel grid $\mathcal{V}_l \in \mathbb{R}^{M \times H \times W}$ that encodes temporal activity at different scales. This yields a unified 4D tensor representation of the event stream as:

\begin{equation}\small~
E_{\text{TPR}} = \left\{ \mathcal{V}_l \right\}_{l=1}^L \in \mathbb{R}^{L \times M \times H \times W},
\end{equation}
which captures both coarse and fine temporal structures across spatial locations.
Next, we apply a lightweight encoder $f_e$ with three convolutional layers and pooling operations, to compress $E_{\text{TPR}}$ into a low-dimensional feature map $x_e$. Formally, this can be expressed as:
\begin{equation}\small~
x_e = f_e(E_{\text{TPR}}) \in \mathbb{R}^{C \times H' \times W'},
\end{equation}
where $C$ is the output channel dimension, and $H'$, $W'$ are the spatial resolutions of the encoded feature.
This representation is designed to be both compact and discriminative, making it suitable for injection into the diffusion model. Finally, $x_e$ is incorporated into each denoising step of the DDIM backbone via cross-attention (see Sec.~\ref{sec:event-guided-diffusino-module}), ensuring that event-derived edge and texture signals consistently guide the restoration process over time.

\vsubsection{Event-Guided Diffusion Module \label{sec:event-guided-diffusino-module}}

To inject structural priors from events into the denoising process, we introduce an Event-Guided Diffusion Module $f_{eg}$ based on the cross-attention mechanism. As shown in Fig.~\ref{fig:3-Methods}, this module conditions the DDIM sampling trajectory on event features $x_e$ extracted from Sec.~\ref{sec:events-guidance}.
Formally, given an intermediate feature map from the U-Net backbone at time step $t$, we perform cross-attention between the U-Net features (as keys $K$ and values $V$) and the event feature $x_e$ (as queries $Q$). Specifically, the attention weights are computed as:
\begin{equation}\small~
\text{Attention}(Q, K, V) = \text{Softmax} \left( (QK^\top) / \sqrt{d} \right) V,
\end{equation}
where $Q = W_q x_e$, $K = W_k x_t$, and $V = W_v x_t$ are linear projections of the query, key, and value inputs, and $d$ is the feature dimension. The result is an event-guided feature modulation that adapts the denoising behavior to spatial edges and contrast regions captured by the event camera.

This cross-attention operation is integrated into selected intermediate layers of the U-Net across all DDIM steps. By doing so, we ensure that event-derived guidance is progressively injected into the generation process, reinforcing structural consistency and reducing semantic ambiguity during denoising. As empirically shown in Sec.~\ref{sec:experiments}, this module is key to preserving fine details and improving generalization in real hazy scenes.

\begin{figure*}
\centering
\includegraphics[width=\linewidth]{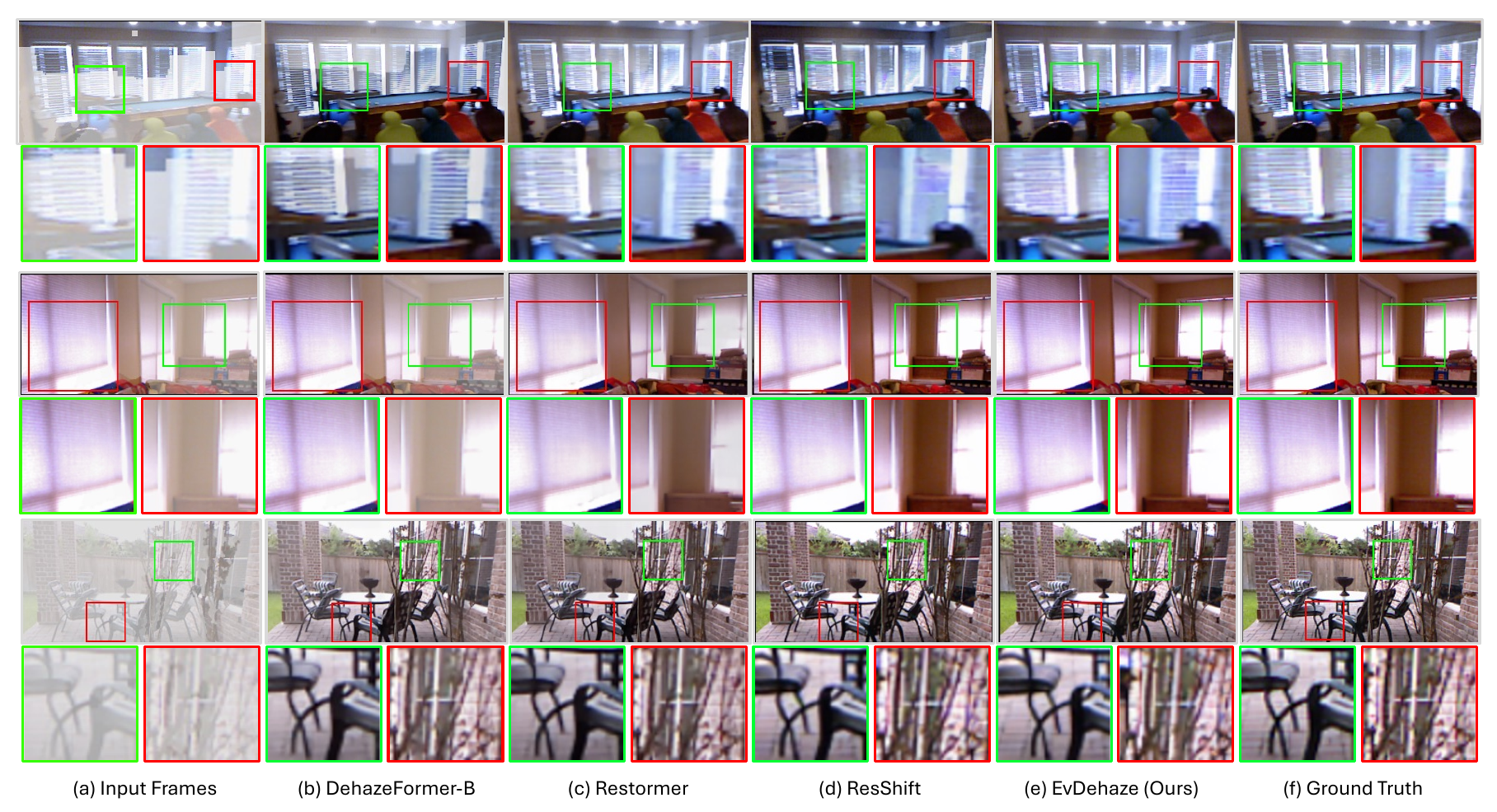}
\vspace{-15pt}
\caption{\small Qualitative comparison on the RESIDE-ITS dataset~\cite{li2018benchmarking}. (a) input frames, (b) DehazeFormer-B, (c) Restormer, (d) ResShift, (e) ours, and (f) ground truth.
Compared with RGB-only methods, EvDehaze recovers sharper edges and more consistent contrast near windows, blinds and outdoor structures, while reducing residual haze and color distortion, and is visually closer to the ground truth.
}
\vspace{-10pt}
\label{fig:visual_compare}
\end{figure*}

\noindent\textbf{Loss Function:}
To train the proposed \textit{EvDehaze} framework, we adopt a combination of pixel-level and perceptual supervision.
Specifically, we optimize the model using the following losses:
\begin{equation}\small~
\mathcal{L}_{\text{total}} = \lambda_{\text{pix}} \mathcal{L}_{\text{pix}} + \lambda_{\text{perc}} \mathcal{L}_{\text{perc}},
\end{equation}
where $\mathcal{L}_{\text{pix}}$ is an $\ell_1$ loss between the generated image $I_o$ and the ground-truth clean image, ensuring accurate color and structure, $\mathcal{L}_{\text{perc}}$ is a perceptual loss \cite{rad2019srobb} computed on deep features extracted from a pretrained VGG network to encourage semantic fidelity.
We empirically set $\lambda_{\text{pix}} = 1.0$ and $\lambda_{\text{perc}} = 0.2$ in all experiments. This simple yet effective combination helps the model recover fine details while maintaining overall perceptual quality.

\vsection{Experiments\label{sec:experiments}}

\noindent\textbf{Impelementation Details:}
The proposed network is implemented using the PyTorch framework and trained on four A800 GPUs. We adopt the AdamW optimizer with a fixed learning rate of $5 \times 10^{-5}$.
For training, we use the Indoor Training Set (ITS) from RESIDE~\cite{li2018benchmarking}, comprising 13,990 hazy images synthesized from 1,399 clean images, along with corresponding simulated event data. For testing, we adopt the indoor split of the Synthetic Objective Testing Set (SOTS), which includes 500 hazy images.
For real-world evaluation with simulated events, we use NH-HAZE~\cite{ancuti2020nh}, which contains 55 hazy-clean pairs, with 50 used for training and 5 for testing. To evaluate performance with real event data, we introduce a newly collected RGB-event dataset captured in haze environments.
During inference, we apply the DDIM sampler with only 15 denoising steps, which balances speed and performance. Other training and architectural settings follow the ResShift pipeline~\cite{yue2023resshift}.
Notably, our event-based conditional guidance module is introduced for the first time in the EvDehaze framework, enabling diffusion-based dehazing guided by high dynamic range events.

\begin{figure*}
\centering
\includegraphics[width=\linewidth]{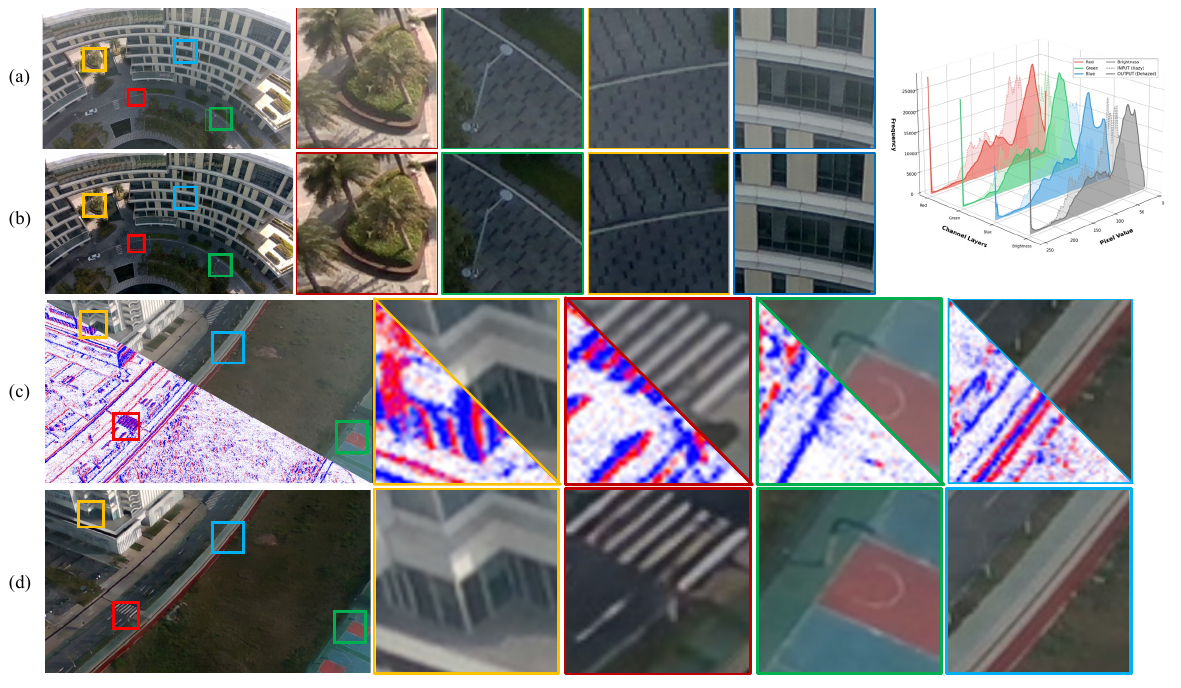}
\vspace{-24pt}
\caption{\small Real-world results on our RGB–event drone dataset under heavy haze.
(a) and (c) are hazy inputs with overlaid event activity, and (b) and (d) are the dehazed outputs of our event-guided diffusion model \textit{EvDehaze}.
Our method removes large-scale haze and recovers clearer edges and textures in distant buildings and roads; the histogram (right) shows an expanded intensity range and improved global contrast compared with the input.}
\vspace{-12pt}
\label{fig:11-Realworld-Evaluation}
\end{figure*}
\begin{figure*}[t!]
\centering
\includegraphics[width=\linewidth]{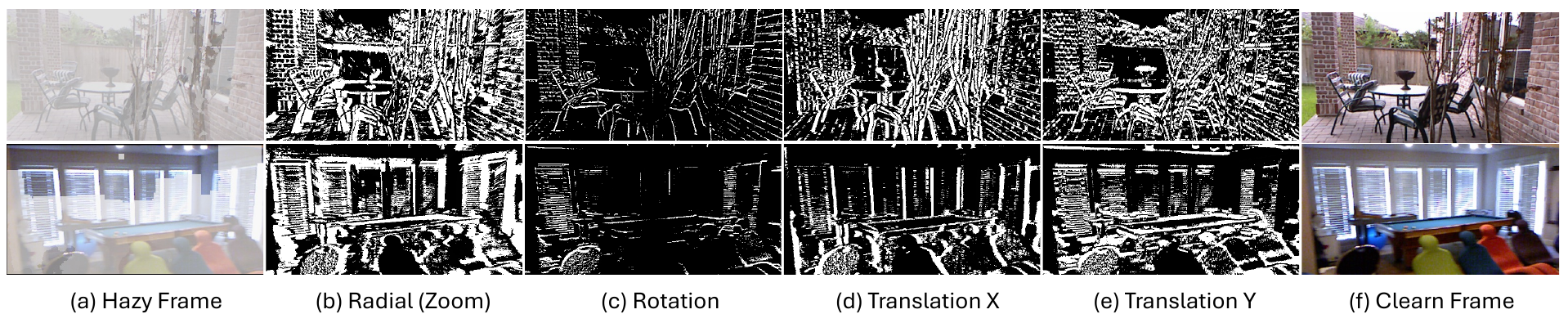}
\vspace{-22pt}
\caption{\small Simulated events under random motions.
Since event polarity is reversible in this visualization, both positive and negative events are shown as white for clarity, while black regions remain static.}
\vspace{-15pt}
\label{fig:9-simulated}
\end{figure*}

\begin{table}[t!]
\setlength{\tabcolsep}{14.6pt}
\renewcommand{\arraystretch}{1.0}
\centering
\caption{\small Ablation studies of the proposed model EvDehaze on SOTS dataset.}
\label{table:abla}
\vspace{-5pt}
\resizebox{\linewidth}{!}{
\begin{tabular}{l|c|c}
\toprule[0.3mm]
EvDehaze Architech  & PSNR$\uparrow$  & SSIM$\uparrow$  \\ \hline
(1) Baseline (ResShift)   & 29.06 & 0.950  \\
(2) w/o event data      & 30.15 & 0.957 \\
(3) w/o cross attention   & 33.65 & 0.981\\
\midrule[0.3mm]
(4) Full Model (EvDehaze) & 34.12  & 0.986 \\
\bottomrule[0.3mm]
\end{tabular}}
\vspace{-20pt}
\end{table}
\noindent\textbf{{Datasets and Evaluation}:}
We assess EvDehaze on three complementary settings:
(i) a \textit{synthetic image} benchmark from dataset RESIDE~\cite{li2018benchmarking} that pairs images with \textit{simulated events}.
(ii) a \textit{real-world image} benchmark from the dataset NH-HAZE~\cite{ancuti2020nh} augmented with \textit{simulated events}.
(iii) a newly collected real-world RGB-event dataset collected in haze environment with 341 AQI, featuring \textit{genuine event data} from the PROPHESSEE EVK4 HD event camera.
For fair competition, we adopt the evaluation setting of Yang et al.~\cite{yang2024unleashing} in (i) and (ii), which normalizes image dimensions across different metho to avoid resolution-induced bias.
Details of each dataset are given in the following section.
Performance is reported with PSNR \cite{hore2010image}, SSIM \cite{wang2004image}, and LPIPS \cite{zhang2018unreasonable}.
\noindent\textbf{\textit{Simulated Dataset}:}\label{sec:sim-data}
To supplement real data, we synthesize events with Vid2E~\cite{gehrig2020video} on the RESIDE-ITS split.
We apply realistic 6-DoF camera motions (vertical, horizontal, forward–backward, and rotation), as shown in Fig. \ref{fig:9-simulated}, to each hazy/clean image pair, generating 44k RGB–event pairs.
These samples provide supervised guidance during training and boost generalization on in-the-wild scenes.
~\textbf{\textit{Real-World Dataset Collection}:}\label{sec:real-dataset}
Because no public set offers synchronized RGB–event pairs in real haze, we build a drone-based platform (Fig.~\ref{1-RealDataset}) equipped with a PROPHESSEE EVK4 HD event camera and an Intel RealSense D435i RGB sensor, aligned by a 3.6 cm 3-D-printed mount.
Flights were conducted over several days in severe pollution (AQI = 314) at 80–120 m altitude, producing 30 min of tightly synchronized data with both circular and linear paths.
RGB frames are recorded at 30 FPS; events are stored in RAW format via the Prophesee SDK. The dataset will be released to encourage further research.

\vsubsection{Comparative Experiments}
\noindent\textbf{Quantitative Comparison:}
Tab.~\ref{tab:dehazing_comparison} reports PSNR, SSIM, and LPIPS on SOTS and NH-HAZE. Within the diffusion group, \textit{EvDehaze} achieves the best overall performance on both datasets and clearly surpasses IR-SDE and ResShift, while using far fewer parameters than IR-SDE.
As expected, supervised CNN/Transformer models obtain higher PSNR on synthetic SOTS because diffusion models typically trade slight distortion loss for better perceptual quality. Therefore, our main comparison focuses on diffusion-based methods, where event guidance brings clear advantages.

To further verify the value of events, we integrate EGDM into the Restormer backbone. This \textit{Restormer+EGDM} variant consistently improves Restormer on both SOTS and NH-HAZE under the same architecture and training setup, with no changes except adding event cues. These gains demonstrate that HDR information from events benefits even a strong transformer baseline, supporting our design of injecting events into diffusion models to enhance both quantitative performance and perceptual realism.
\begin{figure*}
\centering
\includegraphics[width=\linewidth]{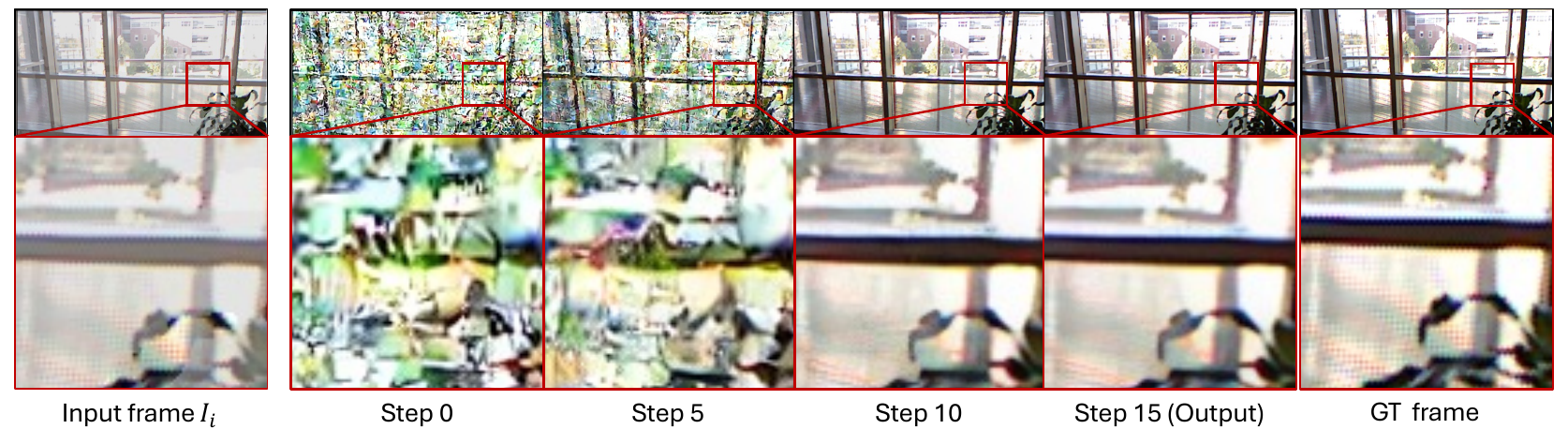}
\vspace{-24pt}
\caption{\small Denoising trajectory of EvDehaze using DDIM sampling. Starting from pure noise, the model restores structural details and contrast, guided by event features. The generated result closely matches the ground truth, showing clear edges and suppressed haze.}
\vspace{-12pt}
\label{4-diffusion_progress}
\end{figure*}
\begin{figure}[t!]
\centering
\includegraphics[width=\linewidth]{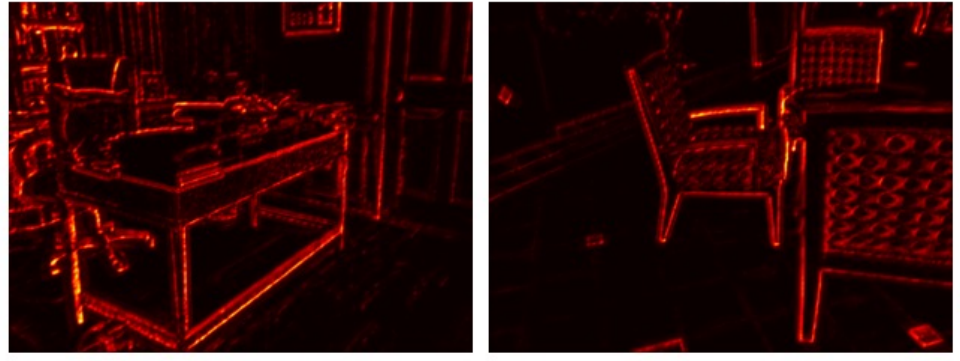}
\vspace{-20pt}
\caption{\small Visualization of the event feature $x_e$ with heatmaps.}
\vspace{-10pt}
\label{fig:event_visual}
\end{figure}

\noindent\textbf{Visualization Comparison:}
Fig.~\ref{fig:visual_compare} presents qualitative results on challenging RESIDE-ITS scenes. Compared with DehazeFormer-B, Restormer, and ResShift, EvDehaze recovers finer structures (window blinds, lamp posts, thin branches) and removes distant haze. Other methods leave residual haze or over-smooth textures, whereas EvDehaze preserves sharp edges and stable contrast, closer to the ground truth. These gains are most visible in high-frequency regions and saturated areas, where PSNR or SSIM may underrepresent perceptual quality.
Fig.~\ref{fig:11-Realworld-Evaluation} shows two real-world cases from our RGB–event drone dataset in heavy haze. Given hazy RGB and events, EvDehaze restores building facades, roads, and court markings in the inputs and avoids strong color shifts. The overlaid event maps show responses on these structures, and the diffusion model uses them as guidance to recover contrast. The input–output histograms show a wider intensity range with stronger mid–high tones, confirming that EvDehaze increases usable dynamic range in real scenes. Overall, these results show that event-guided diffusion yields perceptually more realistic dehazed images, beyond what pixel-wise fidelity alone can capture. \textit{Please refer to the Suppl. Mat. for more visual comparisons.}

\vsubsection{Analytical and Ablation Experiments}

\begin{table}[t!]
\setlength{\tabcolsep}{13.6pt}
\renewcommand{\arraystretch}{1.0}
\centering
\caption{\small Ablation study on different diffusion samplers.}
\label{sample-number}
\vspace{-10pt}
\resizebox{\linewidth}{!}{
\begin{tabular}{l|c|c}
\toprule[0.3mm]
EvDehaze Sampler  & PSNR$\uparrow$  & SSIM$\uparrow$  \\ \hline
(1) DDPM 5 steps   & 33.70 & 0.982 \\
(2) DDPM 15 steps   & 33.92 & 0.985 \\\midrule[0.3mm]
(3) DDIM 15 steps (EvDehaze) & 34.12 & 0.986 \\
\bottomrule[0.3mm]
\end{tabular}}
\vspace{-15pt}
\end{table}

\noindent\textbf{Component Ablation:}
Tab.~\ref{table:abla} evaluates the influence of event guidance and cross-attention.
Removing event data (\textit{w/o event data}) leads to a clear drop in PSNR and SSIM, showing that events supply essential high–dynamic-range cues for restoring contrast under haze.
Removing cross-attention (\textit{w/o cross attention}) also reduces performance compared with the full model, indicating that simple feature concatenation is not sufficient.
These results confirm that both the Events Representation Model and the event-guided cross-attention jointly contribute to the final quality.

\noindent\textbf{Denoising and Sampler Analysis:}
Tab. \ref{sample-number} and Fig.~\ref{4-diffusion_progress} illustrates both the denoising trajectory and the effect of different samplers.
The DDPM sampler improves as its steps increase, while DDIM reaches higher fidelity with only 15 iterations.
The trajectory shows a clear coarse-to-fine evolution: global structure emerges first, followed by progressively refined textures and edges.
This demonstrates that DDIM offers a better efficiency–quality balance for dehazing and highlights a key advantage of diffusion models, gradually restoring details without more artifacts.

\noindent\textbf{Simulated Event Analysis:}
Fig.~\ref{fig:9-simulated} shows simulated events under various motions.
Events reliably respond along boundaries, markings, and texture transitions, while static regions remain inactive.

\noindent\textbf{Event Feature Analysis:}
To illustrate how events guide diffusion, Fig.~\ref{fig:event_visual} visualizes the learned event feature $x_e$ in Fig. \ref{fig:3-Methods}.
The heatmaps highlight stable structural cues, including edges, corners, and texture changes, that persist even under heavy haze and provide reliable guidance to the DDPM.
This confirms that events serve as a strong structural prior that directly shapes the diffusion trajectory.

\vsection{Conclusion}
We present \textit{EvDehaze}, the first event-guided diffusion framework for image dehazing under real-world conditions.
By leveraging the high dynamic range and temporal fidelity of event cameras, our model injects fine-grained structural cues into a latent diffusion process through cross-attention.
Extensive experiments on synthetic and real-world datasets demonstrate not only the superiority of EvDehaze among diffusion-based methods, but also the practical effectiveness of event guidance itself.
In both Restormer+EGDM and EvDehaze, adding event cues consistently improves restoration quality, confirming that events serve as strong structural priors.
\textbf{Limitation:} A current limitation is the reliance on a frozen VQ-VAE backbone, which may restrict adaptability in highly dynamic scenes.

\clearpage

\clearpage
{
\small
\bibliographystyle{ieeenat_fullname}
\bibliography{ref}
}

\end{document}